%% file: main.tex
\documentclass[sigconf]{acmart}

\usepackage{microtype}
\usepackage{graphicx}
\usepackage{subfigure}
\usepackage{hyperref}
\usepackage{multirow}
\usepackage{makecell}
\usepackage{bbding}
\usepackage{algorithm}
\usepackage{algorithmic}
\usepackage{balance}
\usepackage{tablefootnote}
\AtBeginDocument{}

\copyrightyear{2025}
\acmYear{2025}
\setcopyright{acmlicensed}\acmConference[MM '25]{Proceedings of the 33rd ACM International Conference on Multimedia}{October 27--31, 2025}{Dublin, Ireland}
\acmBooktitle{Proceedings of the 33rd ACM International Conference on Multimedia (MM '25), October 27--31, 2025, Dublin, Ireland}
\acmDOI{10.1145/3746027.3755515}
\acmISBN{979-8-4007-2035-2/2025/10}

\acmSubmissionID{4217}
\begin{document}

\title{Learn 3D VQA Better with Active Selection and Reannotation}

\author{Shengli Zhou}
\affiliation{
  \institution{Southern University of Science and Technology}
  \city{Shenzhen}
  \state{Guangdong}
  \country{China}
}
\email{zhousl2022@mail.sustech.edu.cn}
\orcid{0009-0005-2829-990X}

\author{Yang Liu}
\affiliation{
  \institution{Wangxuan Institute of Computer Technology, Peking University}
  \city{Beijing}
  \country{China}
}
\email{yangliu@pku.edu.cn}
\orcid{0000-0002-4259-3882}

\author{Feng Zheng}
\affiliation{
  \institution{Southern University of Science and Technology}
  \institution{Spatialtemporal AI}
  \city{Shenzhen}
  \state{Guangdong}
  \country{China}
}
\email{f.zheng@ieee.org}
\orcid{0000-0002-1701-9141}
\authornote{Corresponding author}

\begin{abstract}
3D Visual Question Answering (3D VQA) is crucial for enabling models to perceive the physical world and perform spatial reasoning. In 3D VQA, the free-form nature of answers often leads to improper annotations that can confuse or mislead models when training on the entire dataset. While other text generation tasks can mitigate this issue by learning on large-scale datasets, the scarcity of 3D scene data enlarges the negative effect of misleading annotations. Although active learning strategies can select valuable instances for training, they fail to identify and resolve misleading labels, which the oracle inevitably provides in practice. To address this issue, we propose a multi-turn interactive active learning strategy. This strategy selects data based on models' semantic uncertainty to form a solid knowledge foundation more effectively and actively requests reannotation from an oracle to resolve potentially misleading labels. For uncertainty assessment, we utilize a variance-based metric that takes semantic relationships between terms into consideration, thus avoiding the uniform inter-class similarity assumption of previous assessment metrics. Extensive experiments exhibit better model performance and a substantial reduction in training costs, with a halving of training costs for achieving relatively high accuracy. The code is available at \url{https://github.com/fz-zsl/AQuA}.
\end{abstract}

%% The code below is generated by the tool at http://dl.acm.org/ccs.cfm. Please copy and paste the code instead of the example below.



\maketitle

\section{Introduction} \label{sec:intro}

3D Visual Question Answering (3D VQA) is a classic multimodal task that involves answering questions based on given 3D scenarios. By training models on the 3D VQA task, we can enhance the model's ability to understand the 3D world and perform spatial reasoning, both of which are crucial for developing embodied agents.

\begin{figure}[t]
    \centering
    \includegraphics[width=0.48\textwidth]{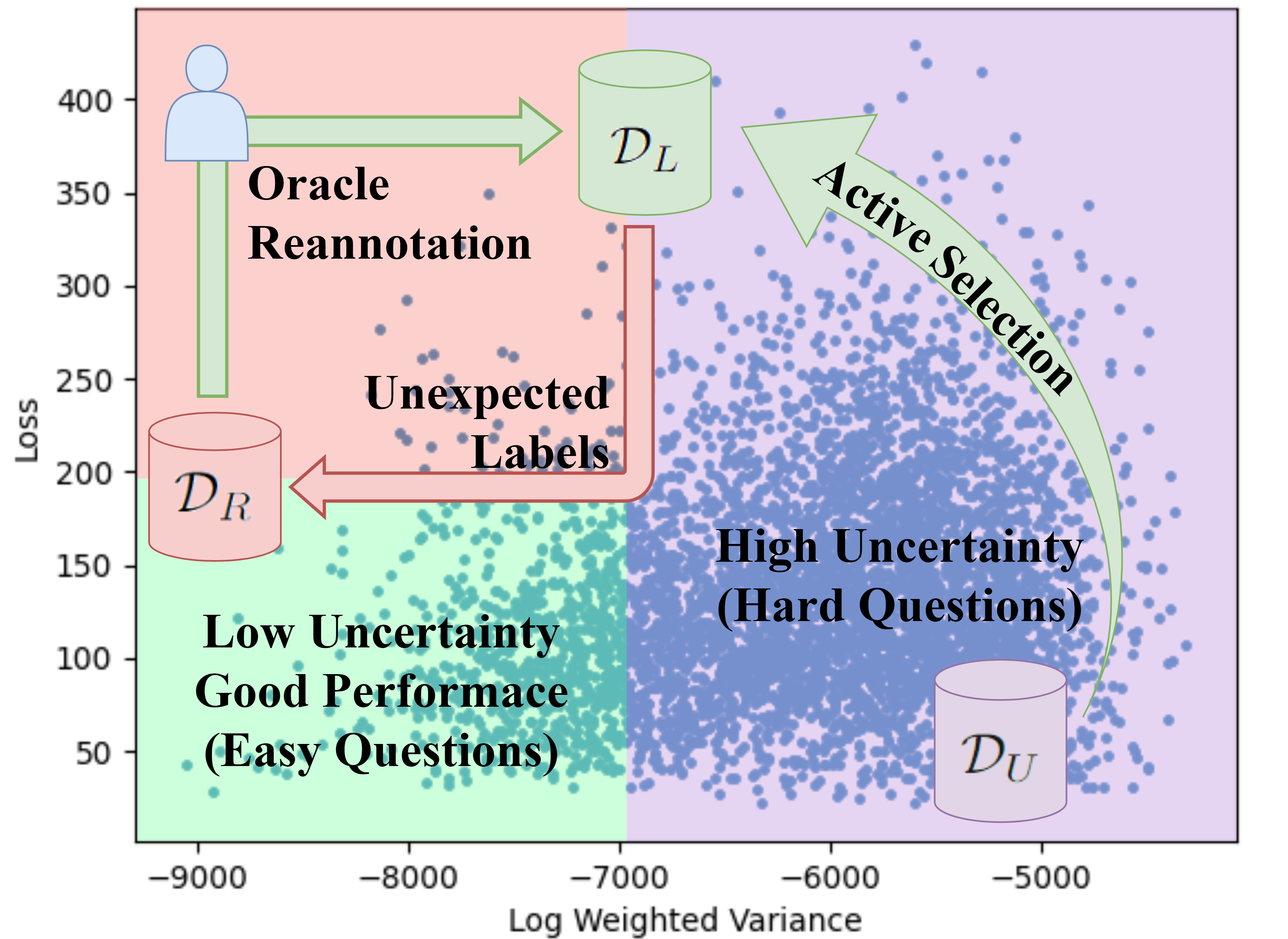}
    \caption{In this paper, we propose a multi-turn interactive active learning strategy. During active selection, the model selects data with the highest uncertainty, which is measured by the weighted variance. Meanwhile, the model also actively requests the oracle to check and reannotate labels that are inconsistent with the model's knowledge. The filtration strategy for making these requests is based on the loss and weighted variance of the predictions.}
    \label{fig:demo}
\end{figure}

In 3D VQA, the answers are of free form, yielding the presence of seemingly correct answers that are indeed improper. For example, answers like ``circle'', ``circular'' and ``round'' may all seem correct to a question about shape, but only ``circle'' is grammatically correct. It is inevitable that some improper answers will appear in the dataset (some improper annotations in the most popular 3D VQA dataset ScanQA \cite{ScanQA} are shown in Appendix B). Though these improper annotations seem acceptable to humans, they can easily confuse or mislead models as they are trained on all data with equal weight. Previously, in other text generation tasks (e.g., Dialogue Generation, Machine Translation, and 2D VQA), models can learn the similarity of different terms by training on large-scale datasets and mitigate the negative impact of inaccurate annotations. However, due to the high cost of collecting 3D scenarios, datasets for 3D VQA are relatively small, magnifying the misleading effect of low-quality annotations. Therefore, a strategy must be applied to extract and utilize valuable data from the training set.

Among machine learning techniques that enable models to self-select training data, active learning stands out as a prevalent strategy. Currently, active learning has been applied to various multimodal tasks \cite{almul_cls,almul_det,almul_con}. Models achieve similar or even better training outcomes with less training cost after employing active learning strategies.

However, in previous active learning strategies, models believe that all labels given by the oracle are correct, and the mechanism of active learning does not have the ability to identify and resolve improper labels inevitably provided by the oracle. Therefore, a superior learning strategy that minimizes the impact of misleading data should be applied to improve the quality of training.

Furthermore, previous active learning strategies for classification are mainly based on the assumption that all classes are distinct and that the differences between each pair of classes are uniform (e.g., \cite{alcls_1,alcls_2}). However, in natural language generation, different classes (i.e., terms or tokens) can exhibit varying degrees of similarity based on semantic relationships, which challenges the initial assumption.

To address these issues, we propose a multi-turn interactive active learning strategy. As shown in Figure \ref{fig:demo}, the strategy consists of two components: active selection and active reannotation. Active selection measures the model's uncertainty by the weighted variance of semantic vectors of predictions. This approach avoids the assumption of uniform inter-class similarity by incorporating semantics into uncertainty calculation. In active reannotation, the model filters out instances that it identifies as inconsistent with other knowledge and requests reannotation from the oracle. By using this framework, we can minimize the impact of misleading data and reduce the cost of fine-grained data cleaning.

In conclusion, our primary contributions are as follows:

\begin{itemize}
    \item We propose a multi-turn interactive active learning strategy with active selection and active reannotation. The method helps 3D VQA models form a better knowledge foundation by selecting valuable data and requesting the reannotation of potentially misleading labels. The strategy enhances the model's learning quality while lowering the cost of fine-grained data cleaning.
    \item We introduced a variance-based method for assessing the model's uncertainty, which takes semantic information into account. This method better adapts to the condition of free-form outputs in 3D VQA.
    \item We validate the effectiveness of active selection, active reannotation, and multi-turn interactive active learning strategy through ablation study and comparative experiments. Extensive results demonstrate the improvements in learning efficiency and performance after applying our methods.
\end{itemize}

\section{Related Work}

\subsection{3D VQA} \label{sec:3dvqa}

3D Visual Question-Answering (3D VQA) is a classic multimodal task that requires models to answer questions based on given 3D scenarios. In 3D VQA, by understanding natural language and reasoning within 3D contexts, models acquire the ability to perceive real physical scenes and reason about spatial information. Furthermore, these capabilities can be leveraged through modules such as Large Language Models (LLMs) to control robots, enabling models to interact directly with real-world environments and respond to questions or instructions \cite{huang2023voxposer}.

Generally, answers in 3D VQA consist of one or several words representing a concept (rather than a full sentence), which can be regarded as a term. Thus, a mainstream method for modeling the 3D VQA task is to treat it as a classification task by employing either a discriminative approach \cite{ScanQA, 3dvista, BridgeQA} or a generative approach \cite{LL3DA, 3DLLM, LEO} to provide answers to questions.

In 3D VQA, due to the high costs of collecting and annotating real-world 3D scenarios, existing 3D scene datasets and 3D VQA datasets are relatively small. For example, the most commonly used 3D VQA dataset, ScanQA, contains only 30 thousand questions, which is much smaller than datasets in other multimodal tasks like 2D VQA. The scarcity of data amplifies the negative impact of improper annotations, leading the model to memorize incorrect patterns. Additionally, since 3D VQA requires a high level of reasoning ability from models, these models tend to have high complexity. As a result, they can easily overfit common patterns in the training set caused by improper annotations. For instance, in ScanQA, answers to $237$ questions about ScanNet scenes 525 to 536 (which accounts for $68\%$) are labeled as ``yes,'' even if the question is not a yes-or-no question. The model may then memorize these patterns, so when it infers about other questions in these scenes, it predominantly predicts ``yes'' as the answer (as detailed in Appendix B). %\ref{sec:bad_anno}

Due to the free-form nature of answers in 3D VQA, improper annotations unavoidably exist in datasets. Also, the scarcity of data strengthens the impact of improper annotations, 3D VQA models require a data selection strategy that can leverage correct and informative data to prevent being misled by low-quality data during training.

\subsection{Active Learning} \label{sec:al}

\begin{figure*}[tp]
    \centering
    \includegraphics[width=.98\textwidth]{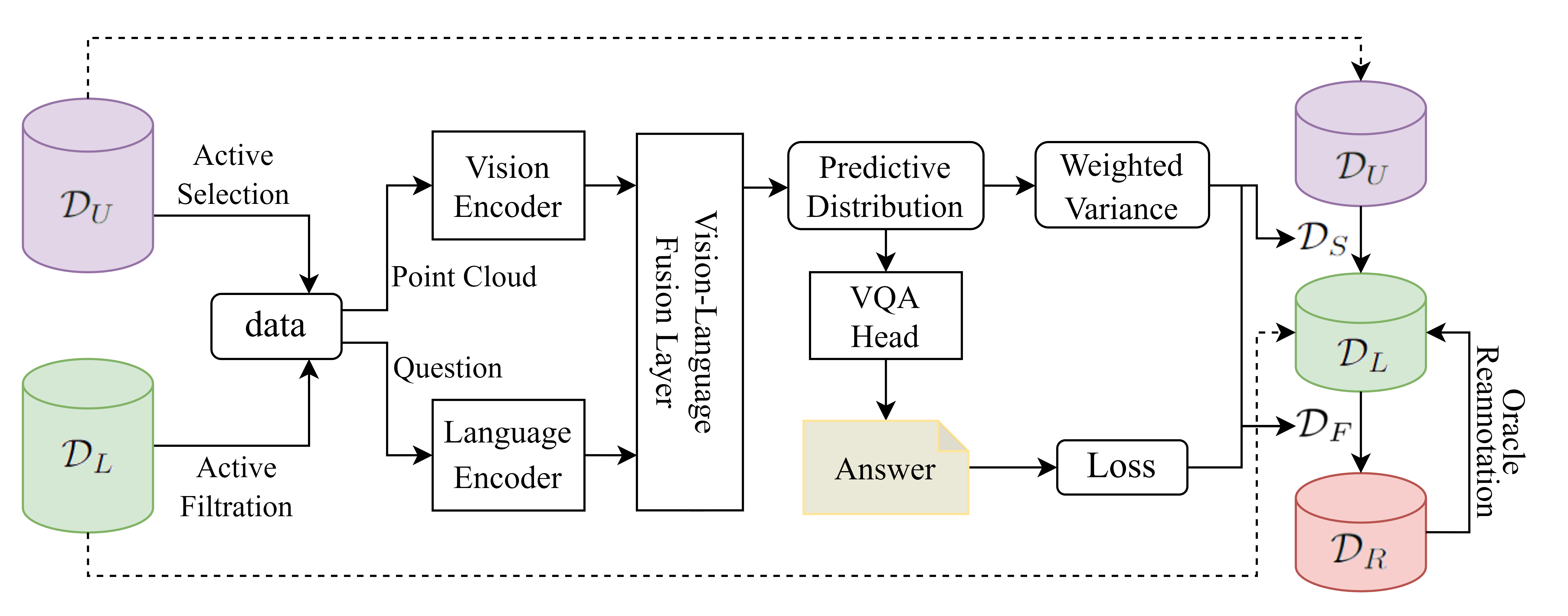}
    \caption{The general workflow of multi-turn interactive active learning strategy. The training set is partitioned into a labeled subset $\mathcal D_L$ and an unlabeled subset $\mathcal D_U$, with all training data initially in $\mathcal D_U$. During active selection, scenes and questions from the unlabeled dataset are fed to the network for inference. The model then calculates the weighted variance of all instances and selects the ones with the highest values to form $\mathcal D_S$. For active reannotation, the model first filters out a subset of the labeled dataset $\mathcal D_F$ by calculating the weighted variance and answer loss of the instances in the labeled dataset. If both values for an instance exceed the thresholds, then the model sends the data to a pool $\mathcal D_R$ and requests the oracle to check and reannotate the instance. Instances are then returned to the labeled dataset after reannotation for further training.}
    \label{fig:workflow}
\end{figure*}

Active learning is a classic machine learning strategy often used to train models on datasets with a large amount of unlabeled data. The core idea of active learning is to allow the model to select data of interest and submit it to an oracle (such as an expert) for labeling, thereby converting this data into labeled data for training. By choosing data that are more important to the model, models trained with active learning strategies are expected to learn to solve the task with fewer labeled data. Moreover, since active learning strategies offer the model with data that it interests most, it can help the model mitigate the issue of unbalanced data distribution, yielding even better model performance.

There are three main types of active learning strategies for data selection: pool-based active learning, stream-based active learning, and active learning with membership queries (also known as query synthesis active learning) \cite{alsurvey}. The first two methods involve selecting data from a fixed unlabeled dataset and a dynamically growing unlabeled dataset, respectively. In methods using membership queries, the model generates samples of interest for the oracle based on prior knowledge about the task. Since generating data that accurately reflects the real distribution is challenging, and the unlabeled dataset is typically fixed, pool-based active learning is the most widely used method.

In active learning, data selection is based on the value of an interest assessment function for each data. There are various ways to design the assessment function. Common approaches include informativeness \cite{LEWIS1994148,Liu_2021_ICCV,BeluchCVPR}, representativeness \cite{alpreclustering,aldiscrrepr,asknlearn}, and expected improvements \cite{Roy2001TowardOA,zhao2021uncertaintyaware,IGSMed}. Based on these strategies, active learning has been applied to multiple Computer Vision tasks, such as detection \cite{Yuan_2023_CVPR,alautodrive,crb}, semantic segmentation \cite{alsemseg,Mackowiak2018CEREALSC,Rangnekar_2023_WACV}, tracking \cite{altrack,ICCVHDDAL}, and action recognition \cite{alactrec}.

In these works, models with active learning exhibit similar or even better performance while using less data. The superiority over traditional training methods indicates that active learning not only helps the model form a solid foundation of prior knowledge for further learning but also mitigates the problem of unbalanced sampling by selecting informative and representative input samples.

However, applying active learning strategies to 3D VQA presents several challenges. First, conventional active learning strategies believe that all labels annotated by the oracle are correct and perfect (which does not hold for 3D VQA datasets), and active learning strategies cannot identify the correctness of labels upon receiving them from the oracle. Hence, data with incorrect or imperfect annotations will continue to deteriorate the model.

Furthermore, conventional assessment functions for classification tasks assume that the similarity between each pair of classes is uniform. However, when models select terms in 3D VQA, the similarity between different terms varies based on their semantics. As a result, previous assessment functions cannot effectively consider the similarity between classes, leading to suboptimal data selection decisions. Therefore, to solve 3D VQA with active learning, a new data selection strategy is required.

\section{Method}

\subsection{General Workflow} \label{sec:workflow}

Previous 3D VQA models typically leverage the entire training set for training. However, due to the inevitable presence of misleading annotations in 3D VQA datasets and the limited scale of 3D VQA datasets, utilizing the entire training set can confuse or mislead the model. Therefore, we need a method that enables the model to autonomously select and learn from valuable data.

In machine learning methods where models self-select data for training, a classic approach is active learning. As described in Section \ref{sec:al}, active learning strategies assess the impact of each datum in the unlabeled dataset $\mathcal D_U$ on the model and select a subset $\mathcal D_S\subseteq\mathcal D_U$ with the highest interest to label and join the labeled dataset $\mathcal D_L$. This process iterates until the annotation budget is exhausted.

While conventional active learning strategies help the model select data that significantly impact its performance, they cannot identify misleading data upon receiving the label from the oracle. To address this issue, a solution is to first use active learning strategies to select data with high uncertainty and train for a certain period. After training, the model evaluates its uncertainty and answer loss for each instance in the labeled dataset $\mathcal D_L$. For instances with low uncertainty and high loss, they indicate that there are unresolvable conflicts between these instances and other data. Such cases may be due to improper annotations in the dataset or insufficient model capability. Hence, the model should request the oracle to check and reannotate these instances. After reannotation, the instances are sent back to the labeled dataset for further training.

In conclusion, the learning framework consists of two active learning operations: active selection and active reannotation. The former (see Section \ref{sec:as}) assesses the model's uncertainty about the answers to questions in $\mathcal D_U$ using weighted variance. It selects the most uncertain questions for the oracle to label and add to $\mathcal D_L$. Active reannotation (see Section \ref{sec:ar}) calculates the weighted variance and loss of all predicted answers in $\mathcal D_L$, filters out instances that are inconsistent with prior knowledge, and requests reannotation from the oracle. The framework is shown in Figure \ref{fig:workflow}, and the pseudo-code is given in Algorithm \ref{alg:framework}.

\begin{algorithm}[!h]
    \caption{Multi-turn Interactive Active Learning Framework}
    \label{alg:framework}
    
    \begin{algorithmic}[1]
        \renewcommand{\algorithmicensure}{\textbf{Define:}}
        \REQUIRE $\Omega_{\text{initial}}$ (the oracle for annotation), $\Omega_\text{reannotation}$ (the oracle for annotation), $k_\text{AS}$ (selection budget for current epoch)
        \ENSURE $\text{cov}_i$ as the weighted covariance (see Section \ref{sec:as}), and $\mathcal L_i$ as the term classification loss (see Section \ref{sec:ar})
        \FOR{epoch\_idx in range(num\_epochs)}
            \IF{require\_selection(epoch\_idx)}
                \STATE $\{\text{cov}_i\}\gets\text{inference}(\mathcal D_U)$
                \STATE $\mathcal D_S\gets\text{selection\_strategy}(\{\text{cov}_i\}, k_\text{AS})$
                \STATE $\mathcal D_L \gets \mathcal D_L \cup \left\{ \Omega_{\text{initial}}(q) \big\vert q \in \mathcal D_S \right\}$
                \STATE $\mathcal D_U\gets\mathcal D_U-\mathcal D_S$
            \ENDIF
            \IF{require\_reannotation(epoch\_idx)}
                \STATE $\{(\text{cov}_i, \mathcal L_i)\}\gets\text{inference}(\mathcal D_L)$
                \STATE $\mathcal D_R\gets\text{filtration\_strategy}(\{(\text{cov}_i, \mathcal L_i)\})$
                \STATE $\mathcal D_L\gets\begin{cases}\Omega_\text{reannotation}(q), &(q, a)\in\mathcal D_R\\(q,a),&\text{otherwise}\end{cases}$
            \ENDIF
            \STATE train($\mathcal D_L$)
        \ENDFOR
    \end{algorithmic}
\end{algorithm}

\subsection{Active Selection} \label{sec:as}

In 3D VQA, the natural language generation task is often modeled as a term selection problem, where the term with the highest probability of being the correct answer is selected from the corpus to be the response.

In previous work, when active learning is applied to traditional classification problems, common strategies for selecting data include high information entropy, high information gain, and large margin values. These strategies, while effective in traditional classification tasks, are not directly applicable to 3D VQA. The main issue is that in traditional classification, any two categories are generally considered mutually exclusive, and the similarity between categories is assumed to be uniform. However, in 3D VQA, the similarity between different terms is determined by semantics. For example, ``rectangle'' and ``rectangular'' should not be treated as completely different categories, and their similarity is different from that between ``rectangle'' and ``triangular''. Hence, when evaluating the model's uncertainty, we need to consider the semantic information of terms.

During model inference, there are two characteristics of the predicted distribution $\mathcal P$ that can indicate a high degree of model uncertainty regarding the answer. The first is the uncertainty in selecting terms, and the second is the uncertainty in the semantics of the term to be selected.

In the first case, the closer the predictive distribution is to the uniform distribution $\mathcal U$, the more uncertain the model is about selecting a term for answering the question. That is, the smaller $D_\text{KL}(\mathcal P||\mathcal U)$, the higher the uncertainty the model has.

In the second case, if the model is certain about the semantics of the answer, the predicted distribution should approximate a Gaussian distribution when projected into the semantic space. Formally, let $\mathcal C$ be the corpus (i.e., the set of all terms) for answers, $p_c$ be the probability of choosing a term $c\in\mathcal C$ as the answer, and $\phi_c$ be the semantic vector of the term $c\in\mathcal C$, then the weighted mean of semantic vectors is defined by

\begin{equation}
    \bar\phi=\sum\limits_{c\in\mathcal C}p_c\phi_c
\end{equation}

and the ideal prediction of distribution should approximate $\mathcal N(\bar\phi, k^{-1}I)$ in the semantic space for some constant $k$.

Thus, in the semantic space, the larger the KL divergence between the predictive distribution and the target distribution $\mathcal N(\bar\phi, k^{-1}I)$, the larger the uncertainty the model has regarding the semantics of the term.

To combine these two factors, we define the uncertainty metric

\begin{equation}
    \Delta=\underbrace{D_\text{KL}(\mathcal P||\mathcal N(\bar\phi, k^{-1}I))}_{\text{Semantic Uncertainty}}-\underbrace{D_\text{KL}(\mathcal P||\mathcal U)}_{\text{Term Confidence}}
\end{equation}

By the definition of KL divergence, we have

\begin{align}
    \Delta&=\sum\limits_{c\in\mathcal C}p_c\log\left(\dfrac{p_c}{\mathcal N(\phi_c|\bar\phi,k^{-1}I)}\right)-\sum\limits_{c\in\mathcal C}p_c\log\left(\dfrac{p_c}{1/|\mathcal C|}\right)\\
    &=\dfrac k2\sum\limits_{c\in\mathcal C}p_c||\phi_c-\bar\phi||_2^2-\dfrac m2\log\left(\dfrac k{2\pi}\right)-\log|\mathcal C|
\end{align}

Proof of the above formula is detailed in Appendix A. %\ref{sec:uiproof}

When selecting data from the unlabeled dataset $\mathcal D_U$ for annotation by the oracle, the model should choose those with high uncertainty, meaning data with large $\Delta$ values. Since $\dfrac m2\log\left(\dfrac k{2\pi}\right)+\log|\mathcal C|$ is a constant, we have

\begin{equation}
    \Delta\propto\sum\limits_{c\in\mathcal C}p_c||\phi_c-\bar\phi||_2^2\triangleq \text{var}_p(\phi)
\end{equation}

Therefore, it is equivalent to selecting data that exhibits high weighted variance in the predictive distribution.

In practice, due to the Curse of Dimensionality, the values for $\text{var}_p(\phi)$ are often very similar across different instances. To address this issue, we alternatively use the logarithm of the determinant of the covariance matrix to rank the instances. Specifically, after each epoch of training, the model performs inference on all instances in $\mathcal D_U$ and predicts the probability $p_c$ of selecting term $c\in\mathcal C$ as the answer. Then, the model calculates $\log|\text{cov}_p(\phi)|$, where $\text{cov}_p(\phi)$ is the weighted covariance matrix of the predicted semantic vectors. Finally, the model selects $k_\text{AS}$ instances with the largest $\log|\text{cov}_p(\phi)|$ values to form $\mathcal D_S$.

\subsection{Active Filtration for Reannotation} \label{sec:ar}

In 3D VQA, since the answers are of free form, datasets inevitably contain improperly labeled data. For example, the widely-used 3D VQA dataset ScanQA includes answers that are irrelevant to the question or in a non-canonical form (e.g., answering ``brown chair'' when the question queries the color). These improper annotations pose a challenge for conventional active learning approaches. In previous active learning strategies, models select the data they find most interesting and utilize these instances for training afterward. Consequently, data with improper labels can continue to mislead the model. To address this issue, we need to implement a mechanism that enables the model to detect and resolve instances with misleading annotations, minimizing the negative impact of such data on the model.

When learning a question-answer pair, the outcome can be categorized into the following three cases:

\textbf{If the model learns the knowledge in the instance effectively}, it will be confident and correct in this instance, resulting in \textbf{low weighted variance and low loss}. In this case, since all the knowledge is compatible and the majority of the dataset is correctly labeled, we can consider the instance to be correctly labeled.

\textbf{If the model does not have sufficient capability to understand the knowledge}, then it will be uncertain about the knowledge and produce predictions with \textbf{a large weighted variance}. An analogy would be how an elementary school student cannot grasp group theory and randomly guess answers in tests. Under this circumstance, the model cannot determine the correctness of the data. Meanwhile, the knowledge in this instance does not contradict the model's prior knowledge (otherwise, the model would yield a certain answer different from the ground truth). Thus, the model can regard this data as correctly labeled and use it for training.

\begin{figure}[t]
    \centering
    \includegraphics[width=0.48\textwidth]{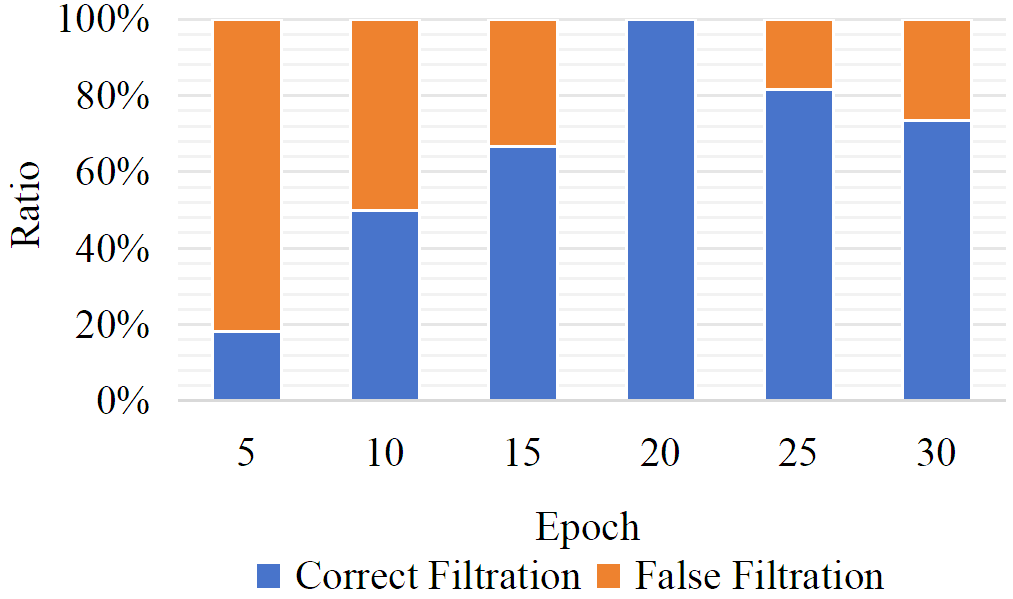}
    \caption{The ratio of instances filtered out due to false filtration and correct filtration across different epochs. At the beginning of training, as the model acquires more knowledge, it becomes better at distinguishing misleading data. After sufficient training, the model achieves its highest accuracy. However, as the model begins to overfit the training data, the accuracy declines, and it starts filtering out instances with less common patterns.}
    \label{fig:exclrate}
\end{figure}

\textbf{When the knowledge is incompatible}, the model is certain of an answer that differs from the ground truth, resulting in \textbf{high loss and low weighted variance}. In this case, the instance is likely to be improperly annotated, but it may also be a difficult instance that the model cannot yet fully comprehend. To reduce the impact of potentially misleading annotations while preserving valuable hard instances, the model can \textbf{\textit{request the oracle to check and reannotate instances with low weighted variance and high loss}}. After reannotation, these instances are returned to the labeled dataset $\mathcal D_L$ for further training.

To validate the correctness of the above statements, we use a trained model to infer on the validation set and calculate both the log determinant of the weighted covariance matrix $\log|\text{cov}_p(\phi)|$ and term classification loss $\mathcal L$ for each instance. The results are presented in the scatter plot shown in Figure \ref{fig:demo}, where the instances are categorized into three types based on the above analysis. A selection of instances corresponding to each category is listed in Appendix D. %\ref{sec:qinpart}

Finally, the aforementioned method relies on the model's prior knowledge to determine the correctness of data. As such, when the model underfits or overfits the training data, it may filter out more instances that are actually correct, resulting in unnecessary reannotation. The ratios of correctly filtered-out improper annotations are shown in Figure \ref{fig:exclrate}, with detailed examples provided in Appendix C. The change in ratio indicates that: while reannotating early helps the model form a better knowledge foundation, reannotating before the model overfits reduces the cost of reannotation. %\ref{sec:fltdata}

\subsection{Hierarchical Reannotation Strategy}\label{sec:hrs}

Generally, answers in 3D VQA can be categorized into four types: (1) a correct and unique answer to the question; (2) one of the viable answers to the question, such as one of the correct descriptions of the position of an object; (3) an answer relevant to the question but in a non-canonical form, such as ``2 chairs'' when asking for the number of chairs or ``rectangular'' for queries about the shape; or (4) an answer irrelevant to the question (i.e., an incorrect answer), such as ``yes'' when asking about the quantity.

In these cases, the first two types of answers are valid for model training, while the latter two, due to their potential to mislead the model, need to be resolved.

\begin{figure}[t]
    \centering
    \includegraphics[width=0.48\textwidth]{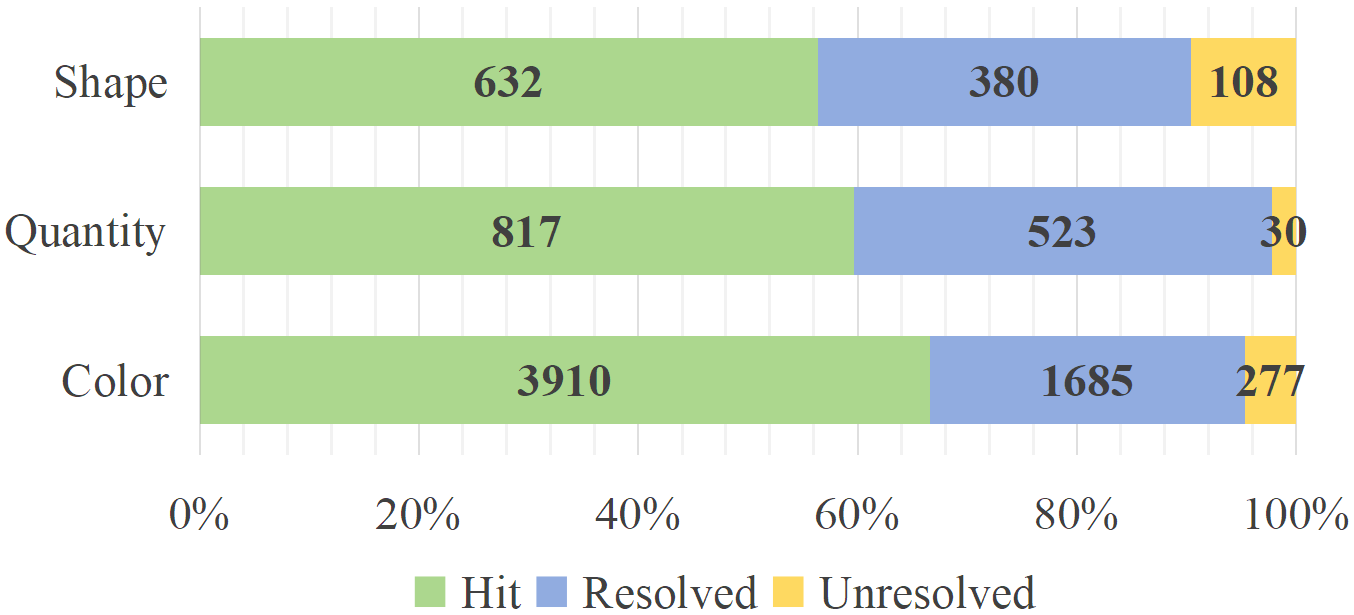}
    \caption{The ratio of different cases in matching. \textbf{Hit} (green) represents answers that are in $\mathcal C'$, indicating a canonical annotation. \textbf{Resolved} (blue) refers to answers that are non-canonical but can be mapped to $\mathcal C'$ using mapping function $\Omega_0:\mathcal C\to\mathcal C'$. \textbf{Unresolved} (yellow) indicates misleading labels that cannot be resolved by the mapping function and require manual reannotation. Since the proportion of unresolved instances is small, the hierarchical reannotation strategy can effectively reduce the cost of reannotation.}
    \label{fig:dataset}
\end{figure}

To filter out data with non-canonical expressions and incorrect answers for data analysis, we first train a model using our active selection strategy and perform inference on all data in training and validation sets. Then, we filter out misleading data with the active filtration strategy and analyze their characteristics.

Through our analysis, answers with non-canonical expressions mostly occur in questions about quantity, color, and shape, while irrelevant answers appear across all types of questions but count for a relatively small proportion.

For non-canonical expressions, we filter concise answers from the corpus and merge terms that convey similar meanings (such as ``red'' and ``reddish'') to create a refined corpus, $\mathcal C'$. During the reannotation of questions related to quantity, color, and shape, the oracle can check for conciseness by matching non-canonical answers with terms in $\mathcal C'$. The heuristic rules for matching are detailed in the supplementary material.

To address matching failures and incorrect annotations, the oracle can perform manual reannotation.

To combine the matching strategy with manual reannotation, we propose a hierarchical reannotation strategy. Since the matching rule is straightforward, a mapping function $\Omega_0:\mathcal C\to\mathcal C'$ is applied first to resolve as many non-canonical labels as possible. The oracle is then only required to handle the remaining unresolvable cases. As shown in Figure \ref{fig:dataset}, the majority of improper annotations can be corrected using the mapping function. Thus, the hierarchical strategy can significantly reduce the cost of manual reannotation.

\section{Experiments}

In this section, we evaluate the advantages of the multi-turn interactive active learning strategy over random-based and conventional active learning strategies. To assess its effectiveness, we apply different active learning approaches to two existing 3D VQA models, ScanQA \cite{ScanQA} and 3D-VisTA \cite{3dvista}, and compare the results to verify the improvements enabled by the multi-turn interactive strategy.

\subsection{Implementation Details}

\subsubsection{Thresholds for Active Filtration}

\begin{figure}[ht]
    \centering
    \subfigure[$\log|\text{cov}_p(\phi)|$]{
        \includegraphics[width=0.27\textwidth]{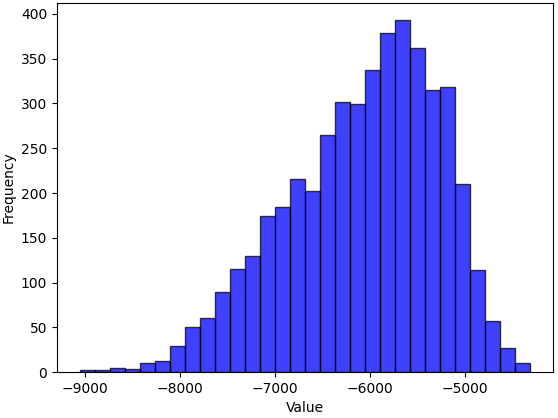}
        \label{fig:hist_variance}
    }\\
    \subfigure[$\mathcal L$]{
        \includegraphics[width=0.27\textwidth]{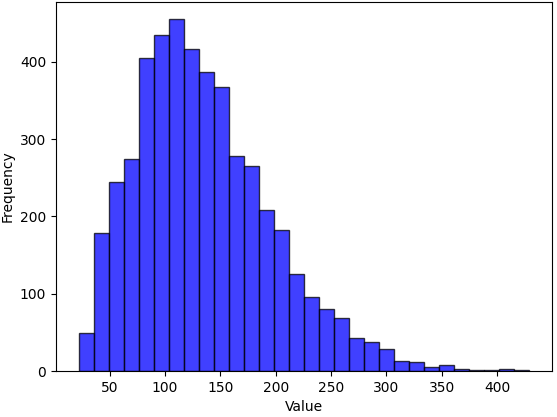}
        \label{fig:hist_loss}
    }
    \caption{Histograms of $\log|\text{cov}_p(\phi)|$ and $\mathcal L$.}
    \label{fig:histograms}
\end{figure}

To categorize all question-answer pairs in $\mathcal D_L$ using weighted variance and loss, thresholds must be set for each metric. During training, we plot the histograms of $\log|\text{cov}_p(\phi)|$ and term classification loss $\mathcal L$ to analyze their distributions. An example histogram is shown in Figure \ref{fig:histograms}. The histograms indicate that both variables follow a skewed distribution. Therefore, we apply Z-score to measure the degree of deviation and determine the thresholds.

After examining the distribution of instances in each category on the scatter plot (i.e., Figure \ref{fig:demo}) and tuning on the validation set, we set the threshold for weighted covariance to $Z_\text{cov}=-1$ and for loss to $Z_\mathcal L=3$. That is, the model will request the oracle to reannotate data in $\mathcal D_L$ that satisfy both $\log|\text{cov}_p(\phi)|<\mu_\text{cov}-\sigma_\text{cov}$ and $\mathcal L>\mu_\mathcal L+3\sigma_\mathcal L$.

\begin{figure}[t]
    \centering
    \includegraphics[width=0.4\textwidth]{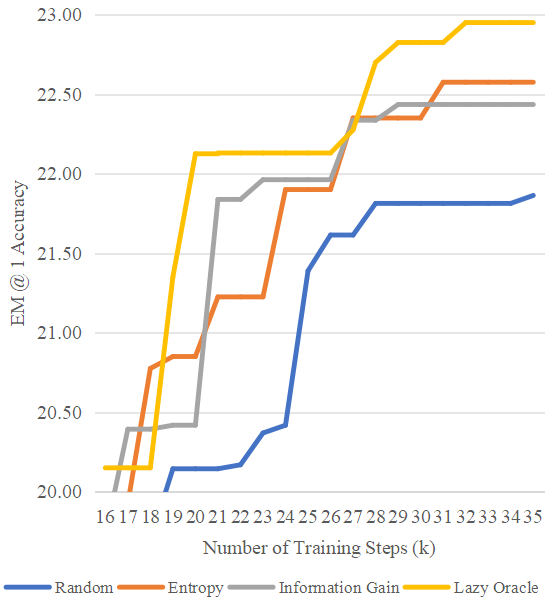}
    \caption{Line graph of ScanQA's best cumulative scores.}
    \label{fig:dalexp}
\end{figure}

\begin{table*}[t]
    \centering
    \caption{The results on the ScanQA dataset from the ``lazy oracle'' experiment show that our active selection strategy outperforms other active learning strategies. This highlights the advantage of considering semantics through weighted variance when selecting informative instances for the model to learn.} \label{tab:lazy}
    \begin{tabular}{cc|cccc|cc}
    \hline
        \multicolumn{2}{c|}{\multirow{2}{*}{\textbf{Model}}} & \multicolumn{4}{c|}{\textbf{ScanQA}} & \multicolumn{2}{c}{\textbf{3D-VisTA}} \\
        ~ & ~ & Random & Entropy & Information Gain & \textbf{Lazy Oracle (Ours)} & Random & \textbf{Lazy Oracle (Ours)} \\ \hline
        \multirow{2}{*}{First 20\% +} & Training Cost & 19 & 18 & 17 & 15 & 13 & 11 \\ 
        ~ & Reduction & 0.00\% & 5.26\% & 10.53\% & 21.05\% & 0.00\% & 15.38\% \\ \hline
        \multirow{2}{*}{First 21\% +} & Training Cost & 25 & 21 & 21 & 19 & 14 & 13 \\ 
        ~ & Reduction & 0.00\% & 16.00\% & 16.00\% & 24.00\% & 0.00\% & 7.14\% \\ \hline
        \multirow{2}{*}{First 22\% +} & Training Cost & 39 & 27 & 27 & 20 & 15 & 14 \\ 
        ~ & Reduction & 0.00\% & 30.77\% & 30.77\% & 48.72\% & 0.00\% & 6.67\% \\ \hline
        \multirow{2}{*}{First 23\% +} & Training Cost & N/A & N/A & N/A & 36 & 20 & 16 \\ 
        ~ & Reduction & N/A & N/A & N/A & N/A & 0.00\% & 20.00\% \\ \hline
        \multirow{2}{*}{AUC} & Area & 0.2056  & 0.3413  & 0.3486  & 0.4211  & 0.5516 & 0.5794 \\ 
        ~ & Improvement & 0.00\% & 65.99\% & 69.58\% & 104.80\% & 0.00\% & 5.04\% \\ \hline
    \end{tabular}
\end{table*}

\begin{table}[t]
    \centering
    \caption{The results on the SQA3D dataset from the ``lazy oracle'' experiment.} \label{tab:lazy_sqa}
    % \vskip 0.15in
    \begin{tabular}{cc|cc}
    \hline
        \multicolumn{2}{c|}{\multirow{2}{*}{\textbf{Model}}} & \multicolumn{2}{c}{\textbf{3D-VisTA}} \\
        ~ & ~ & Random & \textbf{Lazy Oracle (Ours)} \\ \hline
        \multirow{2}{*}{First 45\% +} & Training Cost & 13 & 10 \\ 
        ~ & Reduction & 0.00\% & 23.08\% \\ \hline
        \multirow{2}{*}{First 48\% +} & Training Cost & 27 & 12 \\ 
        ~ & Reduction & 0.00\% & 55.56\% \\ \hline
        \multirow{2}{*}{First 49\% +} & Training Cost & N/A & 12 \\ 
        ~ & Reduction & N/A & N/A \\ \hline
    \end{tabular}
\end{table}

\begin{table*}[t]
    \centering
    \caption{The results of experiments for ablation study, ``AS'' and ``AR'' represent the usage of variance-based active selection and active reannotation, respectively. Results under most metrics show that models gain improvements from both active selection and active reannotation. The results validate the importance of building a strong knowledge foundation through active selection and mitigating the negative impact of misleading data through active reannotation.} \label{tab:dili}
    \begin{tabular}{cccc|ccccccc}
    \hline
        \textbf{Model} & \textbf{Strategy} & \textbf{AS} & \textbf{AR} & \textbf{EM @ 1} & \textbf{BLEU-1} & \textbf{BLEU-2} & \textbf{BLEU-3} & \textbf{BLEU-4} & \textbf{CIDEr} & \textbf{ROUGE-L} \\ \hline
        \multirow{10}{*}{ScanQA} & Random & \XSolidBrush & \XSolidBrush & 19.19  & 28.78  & 18.68  & 13.44  & 9.53  & 59.05  & 31.51  \\ \cline{2-11}
        ~ & Ablation & \XSolidBrush & \CheckmarkBold & 19.59  & 29.52  & 19.22  & 13.79  & 9.11  & 60.11  & 32.01  \\ \cline{2-11}
        ~ & \multicolumn{3}{c|}{\multirow{2}{*}{Improvement}} & 0.40  & 0.74  & 0.54  & 0.35  & -0.42  & 1.06  & 0.50  \\ 
        ~ & ~ & ~ & ~ & 2.10\% & 2.57\% & 2.87\% & 2.64\% & -4.37\% & 1.80\% & 1.60\% \\ \cline{2-11}
        ~ & Lazy Oracle & \CheckmarkBold & \XSolidBrush & 19.83  & 29.03  & 19.49  & 13.69  & 8.26  & 60.37  & 31.81  \\ \cline{2-11}
        ~ & \multicolumn{3}{c|}{\multirow{2}{*}{Improvement}} & 0.64  & 0.25  & 0.81  & 0.25  & -1.26  & 1.33  & 0.31  \\ 
        ~ & ~ & ~ & ~ & 3.34\% & 0.88\% & 4.33\% & 1.87\% & -13.27\% & 2.25\% & 0.97\% \\ \cline{2-11}
        ~ & Diligent Oracle & \CheckmarkBold & \CheckmarkBold & 19.42  & 30.14  & 19.66  & 13.64  & 8.85  & 61.53  & 32.19  \\ \cline{2-11}
        ~ & \multicolumn{3}{c|}{\multirow{2}{*}{Improvement}} & 0.24  & 1.36  & 0.98  & 0.20  & -0.68  & 2.49  & 0.69  \\ 
        ~ & ~ & ~ & ~ & 1.23\% & 4.73\% & 5.24\% & 1.50\% & -7.09\% & 4.21\% & 2.18\% \\ \hline
        \multirow{10}{*}{3D-VisTA} & Random & \XSolidBrush & \XSolidBrush & 22.10  & 30.42  & 21.41  & 16.84  & 12.03  & 67.39  & 34.54  \\ \cline{2-11}
        ~ & Ablation & \XSolidBrush & \CheckmarkBold & 22.16  & 30.56  & 21.29  & 16.64  & 12.16  & 67.58  & 34.91  \\ \cline{2-11}
        ~ & \multicolumn{3}{c|}{\multirow{2}{*}{Improvement}} & 0.06  & 0.14  & -0.12  & -0.20  & 0.13  & 0.19  & 0.37  \\ 
        ~ & ~ & ~ & ~ & 0.29\% & 0.45\% & -0.54\% & -1.18\% & 1.10\% & 0.28\% & 1.06\% \\ \cline{2-11}
        ~ & Lazy Oracle & \CheckmarkBold & \XSolidBrush & 22.40  & 30.85  & 21.76  & 16.82  & 12.73  & 68.31  & 34.98  \\ \cline{2-11}
        ~ & \multicolumn{3}{c|}{\multirow{2}{*}{Improvement}} & 0.30  & 0.42  & 0.35  & -0.01  & 0.70  & 0.92  & 0.44  \\ 
        ~ & ~ & ~ & ~ & 1.36\% & 1.39\% & 1.64\% & -0.08\% & 5.83\% & 1.37\% & 1.27\% \\ \cline{2-11}
        ~ & Diligent Oracle & \CheckmarkBold & \CheckmarkBold & 22.59  & 32.42  & 22.76  & 17.71  & 12.62  & 70.38  & 35.74  \\ \cline{2-11}
        ~ & \multicolumn{3}{c|}{\multirow{2}{*}{Improvement}} & 0.49  & 1.99  & 1.35  & 0.88  & 0.59  & 2.99  & 1.19  \\ 
        ~ & ~ & ~ & ~ & 2.23\% & 6.56\% & 6.32\% & 5.20\% & 4.94\% & 4.44\% & 3.45\% \\ \hline
    \end{tabular}
\end{table*}

\subsubsection{Experimental Settings}

For ScanQA, the model selects $2000$ instances before training and then actively selects $1000$ instances based on the values of $\log|\text{cov}_p(\phi)|$ after each epoch. The active selection process stops when fewer than $100$ instances remain in $\mathcal D_U$. Active reannotation is conducted at epochs $0,5,10,20$, and $25$. We perform active reannotation early in training, as correcting misleading annotations early helps the model build a more solid knowledge foundation.

For 3D-VisTA, since the model has a better capacity to learn new data, we enlarge the size of the data selected in each active selection procedure. Specifically, the model begins with $3000$ instances and actively selects $k_\text{AS}=1500$ instances with the largest $\log|\text{cov}_p(\phi)|$ values after each epoch. However, as $k_\text{AS}$ increases, there are also more valuable data left in $\mathcal D_U$. Thus, to better utilize these data, we gradually decrease $k_\text{AS}$ when $|\mathcal D_U|$ is relatively small. Practically, we adopt

\begin{equation}
    k_\text{AS}=\begin{cases}
    1500&|\mathcal D_U|>2250\\
    0.5|\mathcal D_U|&750<|\mathcal D_U|\leq 2250\\
    0&|\mathcal D_U|\leq 750
    \end{cases}
\end{equation}

The active reannotation strategy is performed at epochs $5,10,15$ and $20$ for 3D-VisTA.

Finally, as our method does not rely on a specific semantic embedding, we utilize a pretrained BERT model to calculate semantic vectors $\{\phi_c\}$ in all experiments.

\subsection{Active Learning Strategy Evaluation}

\subsubsection{Lazy Oracle Experiment}

In this experiment, we aim to evaluate the effect of the active selection strategy. To isolate the influence of active reannotation, we design a ``lazy oracle'' that eliminates manual reannotation and returns all filtered instances back to the labeled dataset without making any modifications. Formally, for any question-answer pair $(q,a)$ in the pool $\mathcal D_R$ for oracle reannotation, the lazy oracle $\Omega_\text{lazy}$ performs the mapping of

\begin{equation}
    \Omega_\text{lazy}: (q,a)\longmapsto (q,a)
\end{equation}

By employing models with random-based selection, as well as strategies based on information entropy, information gain, and weighted variance, we measure the cumulative best score (i.e., exact matching accuracy in percentage) of each model. The results for the ScanQA model on the ScanQA dataset are shown in Figure \ref{fig:dalexp}. For a clearer comparison, we assess the training cost (number of k-training steps for ScanQA and number of training epochs for 3D-VisTA) required for the model to reach a series of fixed score thresholds considering the exact-matching accuracy of the top-confidence answer. Since the scores at the beginning of training exhibit large randomness, we set the score thresholds to several integer percentages close to the model's best performance (namely,  $20\%,21\%,22\%$, and $23\%$ for the ScanQA dataset, and 45\%, 48\%, and 49\% for Situated Question-Answering dataset SQA3D \cite{SQA3D}). Moreover, to evaluate both training cost and model performance using a single metric, we compute the area under the best cumulative score curve for the ScanQA dataset, defined as: $\text{AUC}=\sum_i\max\left\{s_i-20\%,0\right\}$. For a better demonstration, we truncate the curve at $20\%$ accuracy to exclude the initial phase of training, where the scores exhibit high randomness. The results on the ScanQA and SQA3D datasets are presented in Table \ref{tab:lazy} and Table \ref{tab:lazy_sqa}, respectively.

According to the table, applying a non-randomized active learning strategy allows the model to achieve score thresholds with reduced training costs. This improvement benefits from active learning's mechanism of selecting instances that the model interests most. Under such a mechanism, the model can learn from common and basic patterns to harder cases, which helps the model build a strong knowledge foundation at the beginning and span the knowledge space as it is trained on more instances.

Through active learning approaches, the model can not only acquire essential data early on to establish a stronger foundational knowledge for further learning but also mitigate the issue of imbalanced sampling in the dataset. Thus, the line graphs of active-learning-based models also demonstrate an increase in the best model performance compared to baseline models.

Moreover, the weighted variance-based active selection strategy outperforms other non-randomized strategies across all metrics. The results validate that incorporating semantics via weighted variance to assess the model's uncertainty in answers enhances its ability to select informative instances, thereby helping the model more effectively learn the foundational knowledge and hard cases in the 3D VQA task.

\subsubsection{Diligent Oracle Experiment and Ablation Study}

To validate the effectiveness of active reannotation, we construct a model with a randomized selection strategy and an oracle using a hierarchical reannotation strategy (as outlined in Section \ref{sec:hrs}) for the ablation study. We also demonstrate the advantage of applying the multi-turn interactive active learning strategy by a ``diligent oracle'' that utilizes both active selection and active reannotation. In the experiment, we use the random-based active learning strategy without active reannotation as a baseline and assess the performance of the strategies mentioned above. The results for exact-matching accuracy, BLEU, CIDEr, and ROUGE-L, are shown in Table \ref{tab:dili}.

Results in the table verify that both weighted variance-based active selection and active reannotation strategies can individually improve the performance of the model. Moreover, by combining the strategies, our overall multi-turn interactive active learning framework shows greater improvement across most metrics.

Therefore, by detecting and reannotating instances with improper labels, we can effectively prevent misleading annotations from continuously deteriorating the model after being selected. Additionally, the multi-turn interactive active learning framework enables the model to establish a stronger foundational knowledge early in training. The foundation further allows the model to better understand and solve the 3D VQA task, enabling more effective decision-making in both selection and filtration, which ultimately forms a positive feedback loop and benefits the training process.

\section{Conclusion}

In this paper, we propose the multi-turn interactive active learning strategy. The strategy evaluates the model's uncertainty through weighted variance in active selection and resolves improper annotations by requesting reannotation from an oracle. Extensive experiments demonstrate that this approach not only improves the model's performance but also reduces training costs. The results validate that our method enables the model to effectively leverage valuable instances in the dataset, enhancing its ability to span its knowledge space. Furthermore, the multi-turn interactive active learning strategy significantly reduces the cost of fine-grained data cleaning through the hierarchical reannotation strategy, offering an effective solution for training 3D VQA models on noisy datasets.

\begin{acks}
This work was supported by the National Key Research and Development Program of China under Grant 2024YFE0203100 and the China National University Student Innovation and Entrepreneurship Development Program under Grant 202414325005.
\end{acks}

\newpage

\bibliographystyle{ACM-Reference-Format}
\balance
\bibliography{references}

\newpage

\input{appendix}

\end{document}

%% file: appendix.tex
\appendix

\section{Correctness Proof of the Uncertainty Metric} \label{sec:uiproof}

To formulate the uncertainty of the model with respect to input, we define the uncertainty metric

\begin{equation}
    \Delta=\underbrace{D_\text{KL}(\mathcal P||\mathcal N(\bar\phi, k^{-1}I))}_{\text{Semantic Uncertainty}}-\underbrace{D_\text{KL}(\mathcal P||\mathcal U)}_{\text{Term Confidence}}
\end{equation}

Where $D_\text{KL}(p||q)=\sum\limits_ip_i\log\left(\dfrac{p_i}{q_i}\right)$ is the Kullback-Leibler (KL) divergence between distributions $p$ and $q$.

\begin{proposition}
Selecting instances with the largest values for the uncertainty metric $\Delta$ is equivalent to selecting instances with the largest values for the weighted variance $\text{var}_p(\phi)$.
\end{proposition}

\begin{proof}
Let $m$ be the dimension of the semantic space. By the definition of KL divergence, we have:

\begin{align}
    \Delta&=\sum\limits_{c\in\mathcal C}p_c\log\left(\dfrac{p_c}{\mathcal N(\phi_c|\bar\phi,k^{-1}I)}\right)-\sum\limits_{c\in\mathcal C}p_c\log\left(\dfrac{p_c}{1/|\mathcal C|}\right)\\
    &=\sum\limits_{c\in\mathcal C}p_c\log p_c-\sum\limits_{c\in\mathcal C}p_c\log\mathcal N(\phi_c|\bar\phi,k^{-1}I)-\sum\limits_{c\in\mathcal C}p_c\log p_c\\
    &~ ~ ~ ~-\log|\mathcal C|\sum\limits_{c\in\mathcal C}p_c\\
    &=-\sum\limits_{c\in\mathcal C}p_c\log\left\{\left(\dfrac k{2\pi}\right)^{m/2}\exp\left[-\dfrac12(\phi_c-\bar\phi)^T(k^{-1}I)^{-1}(\phi_c-\bar\phi)\right]\right\}\\
    &~ ~ ~ ~-\log|\mathcal C|\\
    &=-\sum\limits_{c\in\mathcal C}\dfrac{mp_c}2\log\left(\dfrac k{2\pi}\right)+\dfrac k2\sum\limits_{c\in\mathcal C}p_c||\phi_c-\bar\phi||_2^2-\log|\mathcal C|\\
    &=\dfrac k2\sum\limits_{c\in\mathcal C}p_c||\phi_c-\bar\phi||_2^2-\dfrac m2\log\left(\dfrac k{2\pi}\right)-\log|\mathcal C|
\end{align}

Since $\dfrac m2\log\left(\dfrac k{2\pi}\right)+\log|\mathcal C|$ is a constant, we have:

\begin{align}
    \Delta&=\dfrac k2\sum\limits_{c\in\mathcal C}p_c||\phi_c-\bar\phi||_2^2+\text{const}\\
    &\propto\sum\limits_{c\in\mathcal C}p_c||\phi_c-\bar\phi||_2^2\\
    &=\text{var}_p(\phi)
\end{align}

Therefore, selecting instances with the largest $\Delta$ values is equivalent to selecting instances with the largest values for the weighted variance $\text{var}_p(\phi)$.
\end{proof}

\section{Improper Annotations in ScanQA and Their Effects} \label{sec:bad_anno}

In ScanQA, improper annotations can generally be classified into two categories: answers that are irrelevant to the question and answers that include non-canonical expressions.

In the first case, the answer is entirely irrelevant to the question. Examples of question-answer pairs in this category are shown in the first four rows of the table below. The example in the last row demonstrates that incorrect patterns can mislead the model.

Notably, for the last three questions, instead of providing reasonable predictions, the model simply predicts ``yes,'' even if the ground truth is correct. Upon further investigation of the dataset, we discovered that for questions related to scenes $525$ to $536$, $68\%$ of the answers are simply ``yes,'' regardless of the actual question. In these scenes, the model finds it optimal to memorize the scene pattern and answer ``yes'' to all questions. Such misleading annotations not only fail to contribute to improving the model's perception and reasoning capabilities but also lead the model to memorize incorrect patterns. Therefore, these questions should be reannotated.

Another case of improper annotations in ScanQA involves non-canonical answers. In this case, the answers provide the correct information but are expressed in an incorrect form. Some examples of such cases are shown in Table \ref{tab:noncan}. In all of these examples, the answers contain redundant expressions, meaning they provide extra, irrelevant information (e.g., ``chairs'' in ``2 chairs''). Additionally, in the first example, while ``rectangular'' conveys the meaning of ``rectangle'', it is grammatically incorrect. Although these answers contain relevant information, they introduce confusion for the model regarding whether to use canonical answers. To eliminate this confusion, these answers should be regularized to a clear and standardized form of the term, helping the model better understand the intended meaning.

\section{Filtered Data during Training} \label{sec:fltdata}

The following table lists the question-answer pairs filtered out in epochs $5,10,15,20,25$, and $30$. ``CA'', ``MA'', ``NE'', and ``WA'' denotes ``correct answer'', ``one of multiple acceptable answers'', ``non-canonical expression'', and ``wrong answer'', respectively. For our analysis, ``CA'' and “MA” are considered correct answers, while ``NE'' and ``WA'' are treated as improper answers.

In the early epochs (e.g., epoch $5$), the model lacks sufficient prior knowledge about the 3D VQA task, and, as a result, it filters out some question-answer pairs that are correctly annotated. As the model trains on more data over time, it becomes better at identifying misleading answers, and the proportion of correct filtration increases. Finally, after epoch $25$, as the model starts to overfit, it becomes less tolerant of less common patterns and ends up filtering out more correct instances.

\section{Examples of Question-Answer Pairs in Each Partition of Scatter Plot} \label{sec:qinpart}

In Figure 1, data points are divided into three categories based on their weighted variance and loss. In this section, we provide examples from each of these categories to further illustrate the distinctions.

\subsection{Easy Questions}

When both weighted variance and loss values are relatively small, the model is confident in its predictions, indicating that these questions are easy to solve. As shown in Table \ref{tab:leftdown}, these questions typically relate to common scenarios or basic attributes of large, easily identifiable objects.

\begin{table}[H]
    \centering
    \caption{Examples of questions with irrelevant answers and model misled by incorrect patterns.} \label{tab:irrelevant}
    % \vskip 0.15in
    \hrule
    \begin{minipage}{0.4\textwidth}
        \centering
        \includegraphics[width=0.9\linewidth]{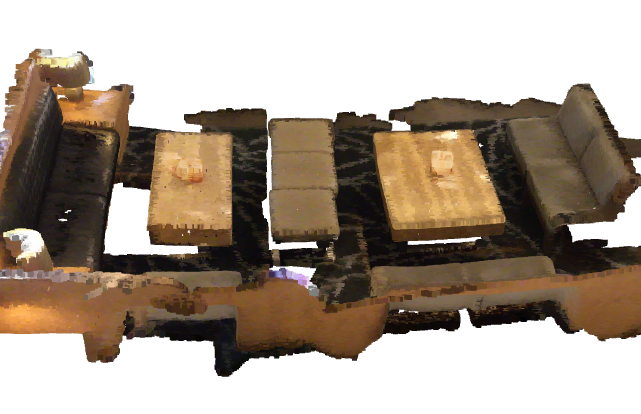}
    \end{minipage}\hfill
    \begin{minipage}{0.5\textwidth}
        Question: How many brown tables is the bench between? \\
        ScanQA's annotation: coffee table \\
        Correct answer: 2 \\
        Model's prediction: 2
    \end{minipage}
    \begin{minipage}{0.08\textwidth}
    ~
    \end{minipage}\hfill
    \hrule
    \begin{minipage}{0.25\textwidth}
        \centering
        \includegraphics[width=\linewidth]{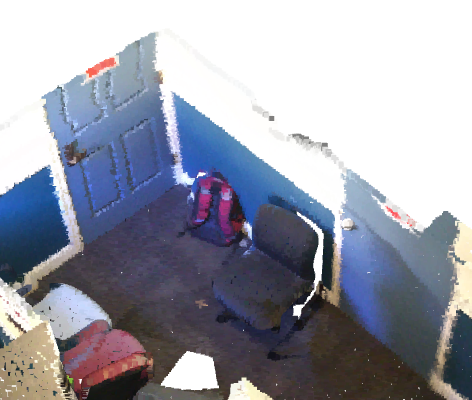}
    \end{minipage}\hfill
    \begin{minipage}{0.22\textwidth}
        Question: What is the color of the door? \\
        ScanQA's annotation: door \\
        Correct answer: white \\
        Model's prediction: white
    \end{minipage}
    
    \hrule
    \begin{minipage}{0.25\textwidth}
        \centering
        \includegraphics[width=\linewidth]{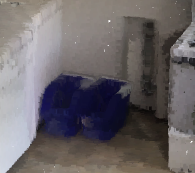}
    \end{minipage}\hfill
    \begin{minipage}{0.22\textwidth}
        Question: What color recycling bin is right of another blue recycling bin? \\
        ScanQA's annotation: yes \\
        Correct answer: blue \\
        Model's prediction: yes
    \end{minipage}
    
    \hrule
    \begin{minipage}{0.25\textwidth}
        \centering
        \includegraphics[width=\linewidth]{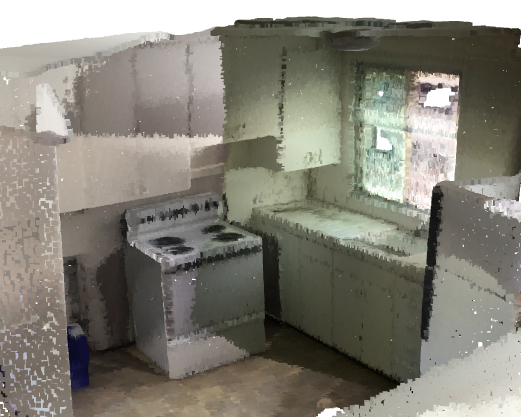}
    \end{minipage}\hfill
    \begin{minipage}{0.22\textwidth}
        Question: What color are the kitchen cabinets? \\
        ScanQA's annotation: 3 \\
        Correct answer: gray \\
        Model's prediction: yes
    \end{minipage}
    
    \hrule
    \begin{minipage}{0.25\textwidth}
        \centering
        \includegraphics[width=\linewidth]{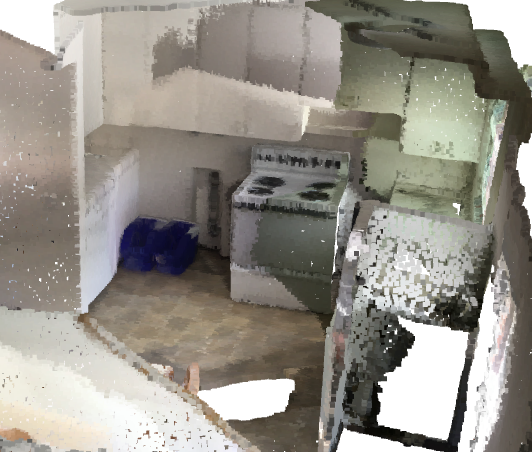}
    \end{minipage}\hfill
    \begin{minipage}{0.22\textwidth}
        Question: What is to the right of the recycling bins? \\
        ScanQA's annotation: stove \\
        Correct answer: stove \\
        Model's prediction: yes
    \end{minipage}
    
    \hrule
\end{table}

\begin{table}[H]
    \centering
    \caption{Examples of questions with non-canonical answers.} \label{tab:noncan}
    % \vskip 0.15in
    \hrule
    \begin{minipage}{0.25\textwidth}
        \centering
        \includegraphics[width=\linewidth]{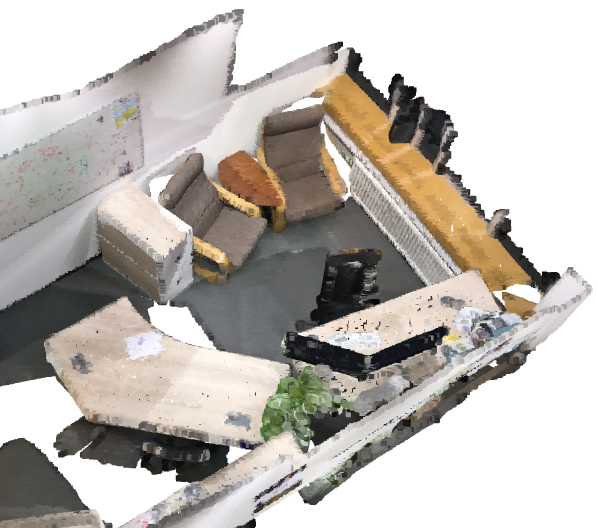}
    \end{minipage}\hfill
    \begin{minipage}{0.22\textwidth}
        Question: What is the shape of the cabinet next to the two chairs? \\
        ScanQA's annotation: rectangular gray cabinet \\
        Canonical answer: rectangle
    \end{minipage}
    
    \hrule
    \begin{minipage}{0.25\textwidth}
        \centering
        \includegraphics[width=\linewidth]{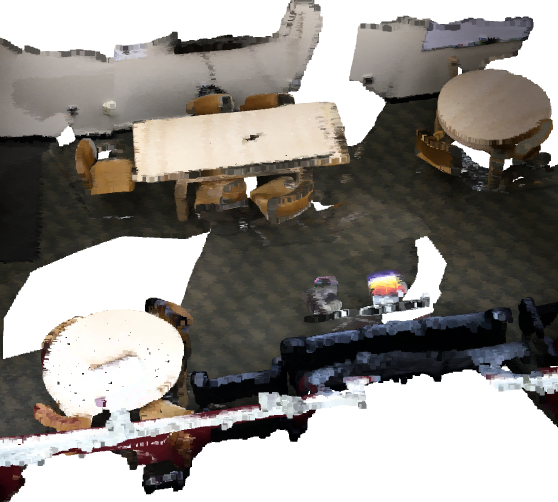}
    \end{minipage}\hfill
    \begin{minipage}{0.22\textwidth}
        Question: What color is the table? \\
        ScanQA's annotation: tan table \\
        Canonical answer: tan
    \end{minipage}
    
    \hrule
    \begin{minipage}{0.25\textwidth}
        \centering
        \includegraphics[width=\linewidth]{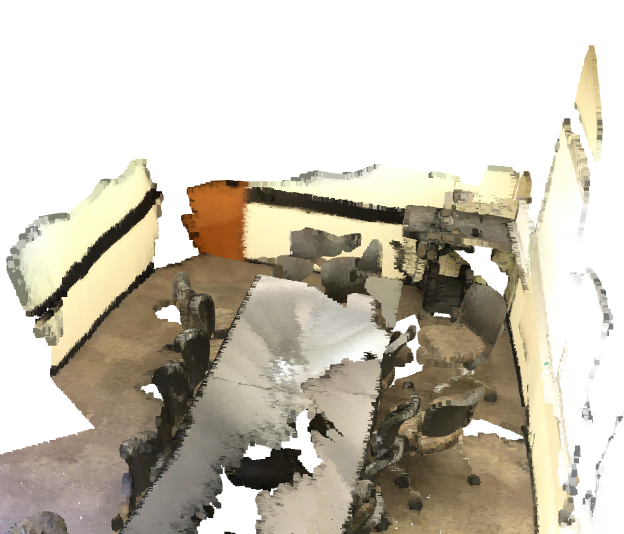}
    \end{minipage}\hfill
    \begin{minipage}{0.22\textwidth}
        Question: What color is the door on the left\tablefootnote{In 3D VQA, the terms ``left'' and ``right'' are conventionally understood from the perspective of an observer at the center of the room.} of the cart? \\
        ScanQA's annotation: brown door \\
        Canonical answer: brown
    \end{minipage}
    
    \hrule
    \begin{minipage}{0.25\textwidth}
        \centering
        \includegraphics[width=\linewidth]{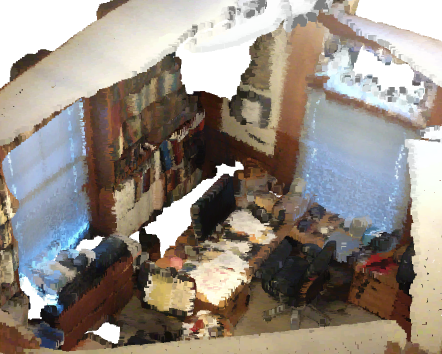}
    \end{minipage}\hfill
    \begin{minipage}{0.22\textwidth}
        Question: What is the monitor near? \\
        ScanQA's annotation: near window \\
        Canonical answer: window
    \end{minipage}
    
    \hrule
    \begin{minipage}{0.25\textwidth}
        \centering
        \includegraphics[width=\linewidth]{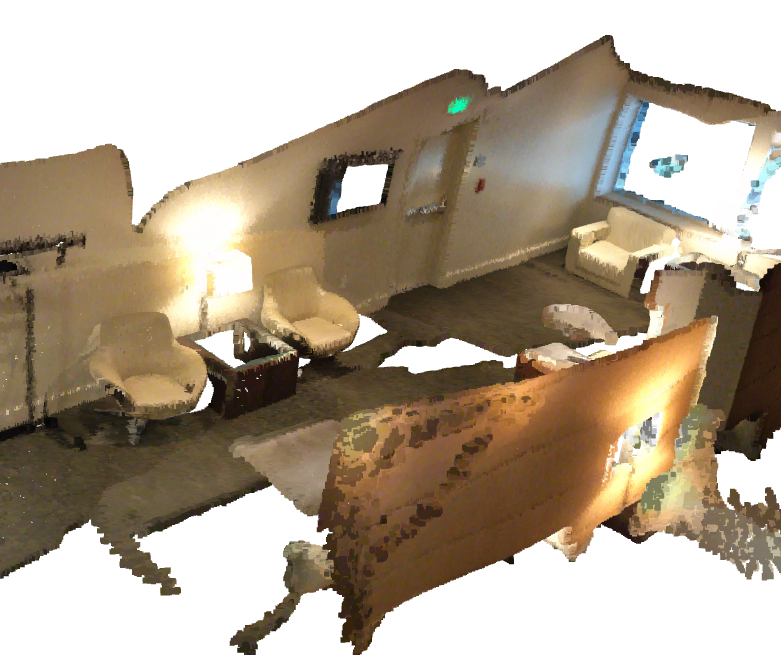}
    \end{minipage}\hfill
    \begin{minipage}{0.22\textwidth}
        Question: How many beige chairs are near the table and lamp? \\
        ScanQA's annotation: 2 chairs \\
        Canonical answer: 2
    \end{minipage}
    \hrule
\end{table}

\begin{table*}[tp]
    \centering
    \caption{Excluded data in different training epochs.}
    % \vskip 0.15in
    \scalebox{0.9}{
        \begin{tabular}{cccc}
        \hline
            \textbf{Epoch} & \textbf{Question} & \textbf{Answer in ScanQA} & \textbf{Note} \\ \hline
            \multirow{11}{*}{5} & Where is the music stand located? & next to chair & MA \\ 
            ~ & What is next to the trash can? & seat cushion & CA \\ 
            ~ & Where is the monitor located? & located on top of desk & NE \\ 
            ~ & What is the shelf above? & above tv stand & MA \\ 
            ~ & Where is the sink located? & on bathroom cabinet & MA \\ 
            ~ & What color are these kitchen cabinets? & light brown & CA \\ 
            ~ & Where is the kitchen cabinet located? & bottom of kitchen counter & MA \\ 
            ~ & What color is the recycling bin? & dark blue & CA \\ 
            ~ & Where is the whiteboard located? & hung on far wall and behind door & MA \\ 
            ~ & Where is the long desk located? & along wall & MA \\ 
            ~ & What is under the lamp? & dark wood nightstand & NE \\ \hline
            \multirow{2}{*}{10} & On what side of the couch is the yellow pillow on? & right side & WA \\ 
            ~ & What color is the chair on the left on the vertical side of a table? & dark blue & CA \\ \hline
            \multirow{3}{*}{15} & What is next to the toilet? & trash bin & CA \\ 
            ~ & What is on the left of a toilet? & toilet paper roll dispenser & WA \\ 
            ~ & How many black trash cans is the picture above? & 2 trash cans & NE \\ \hline 
            20 & What is the color of the door? & door & WA \\ \hline
            \multirow{27}{*}{25} & What shape does the cushion have? & round shape & NE \\ 
            ~ & What shape is the table? & round shape & NE \\ 
            ~ & What color is the towel above the radiator? & maroon & CA \\ 
            ~ & What color base does the lamp have? & gold & WA \\ 
            ~ & What color frame is hung on the wall? & gold & WA \\ 
            ~ & How many windows are there on the right side of the long table? & 2 windows & NE \\ 
            ~ & How many windows are at side of the two long table? & 2 windows & NE \\ 
            ~ & What color are these kitchen cabinets? & light brown & CA \\ 
            ~ & What color are the kitchen cabinets? & brown kitchen cabinets & NE \\ 
            ~ & What color is the dresser? & brown dresser & NE \\ 
            ~ & What color is the table in the center of the room? & orange & WA \\ 
            ~ & What color is the chair at the end of a long wooden table? & light & CA \\ 
            ~ & What shape is the dresser? & dresser & WA \\ 
            ~ & What shape is the brown shelf? & brown shelf & WA \\ 
            ~ & What shape is the top of the table? & round shape table & NE \\ 
            ~ & What color is the tv on the cabinet? & black rectangular tv & NE \\ 
            ~ & How many black trash cans is the picture above? & 2 trash cans & NE \\ 
            ~ & What color are the kitchen cabinets? & 3 & WA \\ 
            ~ & What shape is picture above toilet? & tall rectangular & WA \\ 
            ~ & What shape is the wooden shelf? & rectangular shelf & NE \\
            ~ & What is above the refrigerator? & white kitchen cabinets & CA \\ 
            ~ & What color is the desk on the far left side of the room? & black desk & NE \\ 
            ~ & What color is the chair between a table and a bookshelf? & desk chair & WA \\ 
            ~ & What shape is the black tv? & rectangular tv & NE \\ 
            ~ & What shape is the tv? & rectangular tv & NE \\ 
            ~ & What color is the wooden table set in the middle of the floor? & tan table & NE \\ 
            ~ & What color is the table in the center? & light brown & CA \\ \hline
            \multirow{15}{*}{30} & What color cloth is spread under the brown box? & gold & CA \\ 
            ~ & What are the lamps sitting on? & end tables & CA \\ 
            ~ & What color is the lamp a mix of? & white and gray & WA \\ 
            ~ & What color is the nightstand? & brown night stand & NE \\ 
            ~ & What is the coffee table in front of? & seat & WA \\ 
            ~ & The tv is to the left of what is hung on the wall above the desk? & left of picture & NE \\ 
            ~ & What color is the chair in the middle row? & dark chair & NE \\ 
            ~ & What shape is the coffee table in the middle? & black & WA \\ 
            ~ & What is in the middle of the room? & black table & CA \\ 
            ~ & What is in the center of the wall with a cabinet on each side? & 2 & WA \\ 
            ~ & What is on each side of the closed wooden door? & silver & WA \\ 
            ~ & Where is the small window located? & to right of couch & WA \\ 
            ~ & What sits in front of a couch? & coffee table & CA \\ 
            ~ & What color is the table on the right? & right side & WA \\ 
            ~ & What shape is the curtain to the left of the beds? & black & WA \\ \hline
        \end{tabular}
    }
\end{table*}

\begin{table*}[tp]
    \centering
    \caption{Examples of data with low weighted variance and low loss.} \label{tab:leftdown}
    % \vskip 0.15in
    \begin{tabular}{ccc}
    \hline
        \textbf{Question ID} & \textbf{Question} & \textbf{Answer} \\ \hline
        val-scene0046-56 & What color is the bathtub in the bathroom? & white \\ 
        val-scene0221-101 & What shape is the pillow? & rectangle \\ 
        val-scene0356-17 & What color is the adjustable desk chair? & black \\ 
        val-scene0432-8 & How many sofa chairs face the coffee table? & 2 \\ 
        % val-scene0435-125 & How many beds are in the black cabinet? & 2 \\
        val-scene0435-53 & What color is the bathroom vanity? & brown \\ \hline
    \end{tabular}
\end{table*}

\begin{table*}[tp]
    \centering
    \caption{Examples of data with high weighted variance.} \label{tab:right}
    % \vskip 0.15in
    \begin{tabular}{ccc}
    \hline
        \textbf{Question ID} & \textbf{Question} & \textbf{Answer} \\ \hline
        val-scene0019-72 & What long piece of furniture can seat multiple people? & dark couch \\ 
        val-scene0046-63 & How far from the window curtains is the chair? & few inches from curtains \\ 
        val-scene0063-28 & What does the brown wooden chair lack? & arms \\ 
        val-scene0077-38 & Where is the office printer in the window? & on cabinet \\ 
        val-scene0251-9 & What is the surface of the standing semi-circular wooden object? & flat and smooth \\ \hline
    \end{tabular}
\end{table*}

\begin{table*}[tp]
    \centering
    \caption{Examples of data with low weighted variance and high loss.} \label{tab:leftup}
    % \vskip 0.15in
    \begin{tabular}{cccc}
    \hline % \textbf{\makecell{Original\\Answer}} & \textbf{\makecell{Model's\\Prediction}}
        \textbf{Question ID} & \textbf{Question} & \textbf{GT Label} & \textbf{Prediction} \\ \hline
        val-scene0221-110 & What shape is the beige table? & brown & rectangular \\ 
        val-scene0353-11 & What color is the garment on the other door? & 1 & white \\ 
        val-scene0222-6 & The white pillow is to the left of what color dresser? & to left of desk & white \\ 
        val-scene0222-67 & What color is the door between a bed and the closet? & dark brown door & dark brown \\ 
        val-scene0357-23 & What shape is the shelf next to a flight of stairs? & rectangular shelf & rectangle \\ \hline
    \end{tabular}
\end{table*}

\subsection{Hard Questions}

Question-answer pairs with high weighted variance indicate data where the model is uncertain about its predictions. As shown in Table \ref{tab:right}, these questions are typically phrased in uncommon ways, contain ambiguity, or require the integration of real-world physical knowledge and complex reasoning to derive answers. These are the questions that the model should focus on most during training, which demonstrates the validity of using weighted variance as a measure of model uncertainty.

\subsection{Improper Labels}

As shown in Table \ref{tab:leftup}, improper labels can be classified into two cases. The first case occurs when the answer is irrelevant to the question (e.g., the first three cases in the table). The second case involves answers in non-canonical forms (e.g., the last two cases in the table). In both cases, the loss values are high after training, suggesting that there are conflicts between these instances and the model's knowledge acquired from other data. Additionally, these data points exhibit low weighted variance, indicating that the model is confident in its predictions. Therefore, instances located in the upper-left part of the scatter plot should be checked and reannotated by the oracle.